\pdfoutput=1

\documentclass[11pt]{article}

\usepackage{naacl2021}

\usepackage{times}
\usepackage{latexsym}

\usepackage[T1]{fontenc}

\usepackage[utf8]{inputenc}

\usepackage{microtype}
\usepackage{graphicx}

\usepackage{booktabs}

\usepackage{caption}
\usepackage{subcaption}
\usepackage{scalefnt}

\newcommand\blfootnote[1]{%
  \begingroup
  \renewcommand\thefootnote{}\footnote{#1}%
  \addtocounter{footnote}{-1}%
  \endgroup
}

\newcommand{\benchname}{\textit{SocialAI}~}

\title{SocialAI 0.1: Towards a Benchmark to Stimulate Research on Socio-Cognitive Abilities in Deep Reinforcement Learning Agents}

\author{Grgur Kovač${^{*}}{^{\dagger}}$ \\
        Inria (FR) \\
  
  \And
  
  Rémy Portelas${^{*}}{^{\dagger}}$ \\
  Inria (FR) \\

  \And
  
  Katja Hofmann \\
  Microsoft Research (UK) \\
  
  \And
  
  Pierre-Yves Oudeyer \\
  Inria (FR) \\
  }

\begin{document}

\maketitle

\begin{abstract}
Building embodied autonomous agents capable of participating in social interactions with humans is one of the main challenges in AI. 
This problem motivated many research directions on embodied language use. Current approaches focus on language as a communication tool in very simplified and non diverse social situations: the "naturalness" of language is reduced to the concept of high vocabulary size and variability. In this paper, we argue that aiming towards human-level AI requires a broader set of key social skills: 1) language use in complex and variable social contexts; 2) beyond language, complex embodied communication in multimodal settings within constantly evolving social worlds.
In this work we explain how concepts from cognitive sciences could help AI to draw a roadmap towards human-like intelligence, with a focus on its social dimensions.
We then study the limits of a recent SOTA Deep RL approach when tested on a first grid-world environment from the upcoming \benchname, a benchmark to assess the social skills of Deep RL agents.
Videos and code are available at {\small{\url{https://sites.google.com/view/socialai01}}}.
\end{abstract}

\section{Introduction}
\label{sec:introduction}
\blfootnote{$^{*}$Equal contribution}
\blfootnote{$^{\dagger}$Email grgur.kovac@inria.fr \& remy.portelas@inria.fr}

How do human children manage to reach the social and cognitive complexity of human adults? For Vygotsky, a soviet scholar from the 1920's, a main driver for this path towards "higher-level" cognition are socio-cultural interactions with other human beings \citep{vygot-book}. For him, many high-level cognitive functions a child develops first appear at the social level and then at the individual level. This leap from interpersonal processes to intrapersonal processes is referred to as \textit{internalization}. Vygotsky's theories influenced multiple works within cognitive science \citep{clark-being-there,hutchins96a}, primatology \citep{tomasello-culture-cognition} and the developmental robotics branch of AI \citep{BILLARD98,Brooks2002,colas-imagine}.

A more influential perspective on child development are Jean Piaget's foundational theories of cognitive development \citep{piaget1963}.
For Piaget, the child is a solitary thinker. While he acknowledged that social context can assist development, for him cognitive maturation happens mainly through the child's solitary exploration of their world. The child is a "little scientist" deciding which experiments to perform to challenge its assumptions and improve its representation of the world. 

This Piagetian view on development is well aligned with mainstream Deep Reinforcement Learning (DRL) research, which mainly focuses on sensorimotor development, through navigation and object manipulation problems rather than language based social interactions \citep{dqn,ddpg,imgep,her}. The study of language has been mostly separated from DRL, into the field of Natural Language Processing (NLP), which is mainly focused in learning (disembodied) language models for text comprehension and/or generation (e.g. using large text corpora as in \citet{gpt3}).

In the last few years however, recent advances in both DRL and NLP made the Machine Learning community reconsider experiments with language based interactions \citep{survey-rl-nlp,bender-2020-climbing-NLU}. Text-based exploratory games have been leveraged to study the capacities of autonomous agents to properly navigate through language in abstract worlds \citep{textworld,fantasy-text-world-2020,how-to-dragon-2020}. While these environments allow meaningful abstractions, they neglect the importance of embodiment for language learning, which has long been identified as an essential component for proper language understanding and grounding \citep{cangelosi2010roadmap,Bisk_2020}. Following this view, many works attempted to use DRL to train embodied agents to leverage language, often in the form of language-guided RL agents \citep{chevalierboisvert2018babyai,colas-imagine, Hill2020instrfollow} and embodied visual question answering (EQA) \citep{eqa,DBLP:conf/cvpr/GordonKRRFF18}, and more recently on interactive question production and answering \citep{imitating-intel}.
Multi-agent emergent communication is another subfield which studies how language can emerge from interaction in both embodied and disembodied scenarios \citep{DBLP:conf/aaai/MordatchA18,jacques2019-socialinfl,DBLP:conf/iclr/Lowe0FKP20,DBLP:conf/aaai/WoodwardFH20}.



One criticism that could be made over previous work in light of Vygotsky's theory is the simplicity of the "social interactions" and language-use situations that are considered: in language-conditioned works, the interaction is merely just the agent receiving its goal as natural language within a simple and rigid interaction protocol \citep{survey-rl-nlp}. In Embodied question answering, language-conditioned agents only need to first navigate and then produce simple one or two words answers. And because of the complexity of multi-agent training, studies on emergent communication mostly consider simplistic language (e.g. communication bits). 

In our work, we propose to identify a richer set of socio-cognitive skills than those currently considered in most of the DRL and NLP literature. We organise this set along 3 dimensions: \textit{intertwined multimodality} (coordinating multimodal actions based on multimodal observations), \textit{theory of mind} (inferring other's mental state, e.g. beliefs, desires, emotions, etc) and \textit{social games} (taking part in time-extended structured social interactions).
We then study the failure case of a current SOTA DRL approach on a grid-world social environment.
To enable the design and study of complex social scenarios in reasonable computational time, we consider single-agent learning among scripted agents (a.k.a. Non-Player-Characters or NPCs) and use low-dimensional observation and action spaces. We use templated-language, enabling to emphasize the under-studied challenges of dealing with more natural social and pragmatic situations. 
\paragraph{Social affordances of NPCs.} Although NPCs can be seen as merely complex interactive objects, we argue they are in essence quite different.
NPCs, as humans, can have very complex and changing internal states, including intents, moods, knowledge states, preferences, emotions, etc.
The resulting set of possible interactions with NPCs (social affordances) is essentially different than those with objects (classical affordances). In cognitive science, an affordance refers to what things or events in the environment afford to an organism \cite{social_affordance}.
A flat surface can afford "walking-on" to an agent, while a NPC can afford "obtaining directions from".
The latter is a social affordance, which may require a social system and conventions (e.g. politeness), implying that the NPC must have complex internal states and the ability to reciprocate. Successful interaction might also be conditioned on the NPC's mood, requiring communication adjustments.

Training an agent for such social interactions most likely requires drastically different methods -- e.g. different architectural biases -- than classical object-manipulation training.
We argue that studying isolated social scenarios featuring NPCs in tractable environments is a promising direction towards designing proficient social agents.


\paragraph{Grounding language in social interactions.}
In AI, \textit{natural language} often refers to the ability of an agent to use a large vocabulary and complex grammar. We argue that this is but one dimension of the \textit{naturalness} of language.
Another, often overlooked, dimension of this \textit{naturalness} refers to language grounding, i.e. the ability of an agent to map specific meaning from some domain to language \cite{steels2007symbolgrounding}.
Command following \cite{chevalierboisvert2018babyai,colas-imagine} is an example of language grounding in the environment.
To understand the meaning of "grow green plant", an agent must relate both the plant in the environment to the word "plant", and the word "grow" to the action of watering the plant.
We aim to go a step further by grounding language in social interactions, i.e. requiring social context to be understood in order to make sense of a given utterance. For example, the meaning of a NPC's utterance can change if one knows this NPC is a liar.

\paragraph{Social skills for socially competent agents}
Social skills have been extensively studied in cognitive science \cite{social-skills-assessment1986,social-skills-framework2010} and social and developmental robotics \cite{cangelosi2010roadmap}.
Here we outline some of those skills for the purpose of studying them in the context of training  \textit{social} artificial agents.

\vspace{0.2cm}\textit{\textbf{1 - Intertwinded multimodality~~}} refers to the ability to interact using multiple modalities (verbal and non-verbal) in a coordinated manner. A proficient agent should be able to act using both primitive actions (moving) and language actions (speaking), and to process both visual and language observations (spoken by other NPCs). Importantly, this agent must be able to learn and adapt its multimodal interaction sequence, rather than following a pre-established interaction protocol, e.g. as in EQA. \cite{eqa}, where 1) a question is given to the agent at the beginning of the episode, 2) the agent moves through the environment to gather information, and 3) upon finding an answer it responds (in language) and the episode ends. By the term \textit{intertwined} multimodality we aim to emphasize that the modalities often interchange and the question of "when to use which modality" is non-trivial, e.g. sometimes the relevant information can be obtained by \textit{asking} for it and sometimes by \textit{looking} for it.

    
    
\vspace{0.2cm}\textit{\textbf{2 - Theory of Mind(ToM)~~}} refers to the ability of an agent to attribute to others and itself mental states, including beliefs, intents, desires, emotions and knowledge \cite{wellman1992childToM,flavell1999ToM}.

An agent that has ToM perceives other participants as \textit{minds} like itself.
This enables the agent to theorise about other's intents, knowledge, lack of knowledge etc.
Here we outline some, of many, different perspectives of ToM to better demonstrate how ToM is essential for human social interactions.
\begin{itemize}
    \vspace{-0.15cm}\item \textbf{inferring intents:} the agent is able to infer, based on verbal or non-verbal cues, what others will do or want to do, e.g. that some social peers are liars/trustworthy.
    \vspace{-0.15cm}\item \textbf{false belief:} the agent understands that someone's belief (including its own) can be faulty \cite{bailla2010falsebeliefinfants}.
    \vspace{-0.15cm}\item \textbf{self-awareness:} the capacity to take oneself as the object of thought \cite{objective-self-awareness}.
    \vspace{-0.15cm}\item \textbf{imitating or emulating social peer's behaviour:} agent can imitate a behaviour seen in a social peer, or emulate its goal, e.g. upon observing a peer cut onions the agent is able to cut the onions himself, either with the same movement or with its own strategy.
\end{itemize}
    
    \vspace{0.2cm}\textit{\textbf{ 3 - Social games ~~}} is a concept closely related to pragmatic frames \citep{bruner85pragframes,pragmatic-frames} and language games \citep{wittgenstein1953philosophical}.
    A social game refers to the pattern characterizing the unfolding of possible interactions (equivalent to a "grammar" for social interactions or an interaction protocol). For example, by playing turn taking games a child extracts the rule of each participant having his "turn". It then generalizes this role to a conversation where it understands that it shouldn't speak while someone else is speaking.
    
    Closely related is the concept of roles which was proposed to be one of the key differences between human and ape socio-cognitive abilities \citep{Tomasello-role-of-roles}.
    A human understands that a shared goal is completed by various participants playing different roles, which are often equally important. Crucially, we learn about others' roles by playing our own. 
    For example, in the game of catch where one participant throws the ball and another one catches it. By playing the \textit{catcher} role we understand what the \textit{thrower} role consists of. This makes it easy for us to switch roles and play the \textit{thrower} role.
    
    
    Furthermore, humans have the ability to quickly detect when the \textit{social game} changes and adapt to that change (ex. while playing football we are able to participate in \textit{small talk} with another player).
    

\paragraph{Main contributions:}
\begin{itemize}
\item An outline of the core socio-cognitive skills necessary to enable artificial agents to efficiently act and learn in a social world.
\item A case-study of a SOTA Deep RL approach on a grid-world environment\footnote{Based on Minigrid \cite{gym_minigrid}} featuring scripted NPCs to easily assess social skills.
\end{itemize}

\begin{figure}[htb!]
\centering
\includegraphics[width=0.90\columnwidth]{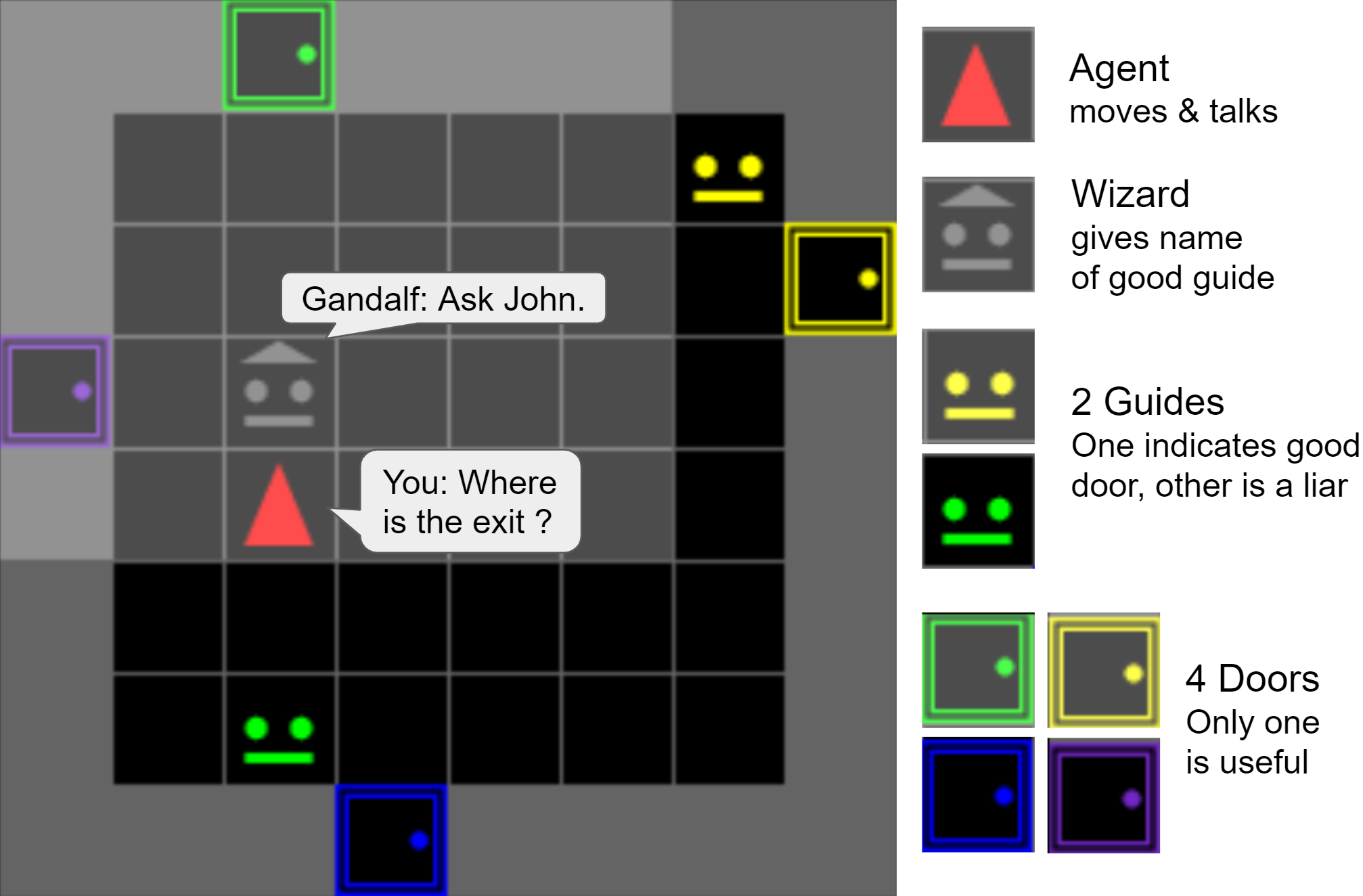}
 \caption{\textit{TalkItOut}, a simple environment to study social skills of DRL agents. Solving it requires to master \textit{intertwined multimodality}, basic \textit{Theory of Mind} (detecting trustworthy agents), and a basic form of \textit{Social Game} (standing near NPCs to interact with them). See app. \ref{app:talkitout_social_skills} for a discussion of required social skills.}
 \label{fig:env}
\end{figure}





\clearpage
\section{Experiments and Results}

Prior to a broader study of social skills in DRL, this work's experiments focus on a simple environment requiring a limited subset of social skills.

\paragraph{TalkItOut}is a one-room grid-world environment. The agent is rewarded upon exiting the room, i.e. saying the right passphrase ("Open sesame") in front of the correct door (out of four, randomly chosen for each new episode). It can both navigate (turn left/right, go forward) and use natural language (template based, 64 possibilities), and observe a partial agent-centric symbolic pixel grid along with the history of observed language outputs from nearby NPCs. To locate the target door, the agent can question three randomly placed NPCs (by asking "Where is the exit" while standing near them): two guides -- one trustworthy and one lying -- and a Wizard that indicates which guide is trustworthy (e.g. "Ask Jack").  NPC names are added to their utterances to allow identification (e.g. "Jack: Go to red door"). See app. \ref{app:env-details} for details, fig. \ref{fig:env} for a visualization.

\paragraph{Implemented Baselines.} Our main baseline is a PPO-trained \cite{ppo} Deep RL architecture
proposed in \cite{hui2020babyai}. We chose this model as it was designed for language-conditioned navigation in grid worlds, which is similar to our setup (although in our case language input is not fixed but varies along interactions). We modify the original architecture to be Multi-Headed (\textit{MH-BabyAI}), since our agent has to both navigate and talk. 
We also consider an ablated version that does not receive language inputs (\textit{Deaf-MH-BabyAI}) and a randomly acting agent (\textit{Random}).
See appendix \ref{app:cond-details} for details.


\paragraph{Results.}
Table \ref{tab:results} shows post-training success rates on a fixed random test-set of $1000$ environments for all conditions. See appendix \ref{app:add-results} for additional results.

The MH-BabyAI agent doesn't solve TalkItOut: its average success rate of 26\% is not statistically significant from Deaf-MH-BabyAI (p>0.05, using Welch's student t-test). MH-BabyAI does not leverage language inputs: both approaches learn the suboptimal policy of going to a random door and saying the passphrase (NPCs are ignored). Similar results are observed on an ablated version of the environment that does not feature the lying guide.

One potential explanation for this failure could be that the language space is too large for the agent. To diagnose whether this is the case, we augment the MH-BabyAI baseline with intrinsic episodic exploration bonuses on observed language (i.e. a curiosity bias). The resulting MH-BabyAI with Exploration Bonuses (MH-BabyAI-EB) manages to solve the ablated no-liar NPC environment, reaching over $0.99$ success rate. However, MH-BabyAI-EB still does not solve the original environment. 

Additional architectural changes were tested as an attempt to improve performances (e.g. language processing, see app. \ref{app:add-results}), without success.

These results showcase that the original TalkItOut environment seems to be a complex challenge DRL learners, especially for the social skills it requires, i.e. handling multi-NPCs multimodal interactions (asking the wizard then the guide) and inferring ill intentions of the false guide.


\begin{table}
\centering
\scalefont{0.94}
\begin{tabular}{@{}lll@{}}
\toprule
Condition \textbackslash~Env. & Original            & No liar NPC \\ \midrule
MH-BabyAI-EB                  & $0.236\pm0.01$   & $\mathbf{0.996\pm0.002}$ \\
MH-BabyAI                     & $0.259\pm0.01$  & $0.252\pm0.010$    \\
Deaf-MH-BabyAI                & $0.260\pm0.02$  & $0.246\pm0.014$ \\ 
Random                        & $0.002\pm1\mathrm{e}^{-3}$  & $0.005\pm0.002$    \\
\bottomrule
\end{tabular}
\caption{Success rate of studied baselines on TalkItOut and a variant without the lying guide (16 seeds, mean $\pm$  stddev, 28M steps). SOTA fails to solve TalkItOut.}
\label{tab:results}
\end{table}


\section{Conclusion And Discussion}

In this work we classified and described the main socio-cognitive skills needed to build socially competent autonomous agents. As a first step towards building \benchname -- a test-bed to assess the social skills of DRL learners -- we performed a preliminary case study on a simple social environment, and showed that a current SOTA DRL approach was unable to learn the required social skills to solve it, making it a relevant test-bed. In future work we plan to release the full \benchname benchmark, which will include more SOTA baselines and multiple complex social environments to encompass a broader range of social skills.

This work suggests that architectural improvements are needed for DRL agents to learn to behave appropriately in multimodal social environments. One avenue towards this is to endow agents with mechanisms enabling to learn models of others' minds, which has been identified in cognitive neuroscience works as a key ingredient of human social proficiency \citep{critiqueRLsociallearning2020}.





\clearpage

\bibliography{anthology,custom}
\bibliographystyle{acl_natbib}

\clearpage
\appendix

\section{Experimental details}

\subsection{Environment}
\label{app:env-details}

\subsubsection{Action space}
The action space of the environment consists of two modalities (\textit{primitive actions} and \textit{language}) which results in a 3D discrete action vector.

The first dimension corresponds to the primitive actions modality. It consists of 7 actions (turn left, turn right, move forward, pickup, drop, toggle, done).
In the TalkItOut task \textit{pickup} and \textit{drop} actions do not do anything and \textit{toggle} and \textit{done} terminate the episode with 0 reward. The reason for this is that we intend to use those actions in the full benchmark.

The second and third dimensions regard the language modality.
The second dimension selects a template (four possibilities) and the third a noun (8 possibilities). The full grammar is shown in table \ref{tab:grammar} 

Both modalities can also be undefined, in which case no action is taken in the undefined modality. Examples of such actions are shown in table \ref{tab:action_examples}.

\begin{table}
\centering
\begin{tabular}{@{}lll@{}}

\toprule
\textbf{Templates} \\
\toprule 
Action & Template \\
\midrule
0 & Where is <noun>. \\
1 & Open <noun>. \\
2 & Close <noun>. \\
3 & What is <noun>. \\

\toprule
\textbf{Nouns} \\
\toprule
Action & Noun \\
\midrule
0 & sesame \\
1 & the exit \\
2 & the wall \\
3 & the floor \\
4 & the ceiling \\
5 & the window \\
6 & the entrance \\
7 & the closet \\
8 & the drawer \\
9 & the fridge \\
10 & oven \\
11 & the lamp \\
12 & the trash can \\
13 & the chair \\
14 & the bed \\
15 & the sofa \\

\end{tabular}
\caption{Template based grammar}
\label{tab:grammar}
\end{table}

\begin{table}
\centering
\begin{tabular}{@{}lll@{}}
\toprule
Action & description \\ \midrule
(1, -, -) &  moves left without speaking   \\
(1, 1, 5) & moves left and utters "Open the window" \\
(-, 1, 5) & doesn't move but utters "Open the window" \\
(-, -, -) & nothing happens \\
\bottomrule
\end{tabular}
\caption{Examples of various actions in the environment. Second and third dimension must both either be underfined or not.}
\label{tab:action_examples}
\end{table}

\subsubsection{State space}

The multimodal state space consists of the \textit{vision} modality and the \textit{language} modality. 

The \textit{vision} modality is manifested as a \textit{7x7} grid displaying the space in front of the agent (shown as highlighted grids in figure \ref{fig:env}). 
Each location of this grid is encoded as three integers  depicting the object type, color and additional information (ex. NPC type: wizard or guide). For example, a blue wizard will be encoded as $(11, 2, 0)$ and a blue guide as $(11,2,1)$. 

The \textit{language} modality is represented as a string containing the currently heard utterances, i.e. utterances uttered by NPCs next to the agent, and their names (ex. "John: go to the green door"). In case of silence an "empty indicator" symbol is used.

As it is often more convenient to concatenate all the utterances heard, to simplify the implementation of the agent, the implementation of the environment also supports giving the full history of heard utterances with the "empty indicator" symbols removed as additional information. 



\subsubsection{The task}

As discussed in the main text the task consists of three NPCs and four doors.
The agent has to find out which door is the correct one by asking the true guide. 
To find out which guide is the correct one the agent has to ask the wizard.
Upon finding out which door is the correct one the agent has to stand in front of it and utter "Open sesame".
Then the episode ends and the reward is calculated by the following equation:

\begin{equation}
r_{extr} = 1.0 - 0.9 * \frac{t}{t_{max}}
\label{eq:extr_reward}
\end{equation}

, where $t$ is the number of steps agent made in the environment and $t_{max}=40$ is the maximum allowed number of steps.
If the agent executes \textit{done}, \textit{toggle} or utters "Open sesame" in front of the wrong door the episode ends with no reward.

An example of a dialogue that might appear in a successful episode is shown in table \ref{tab:successful_episode}
\begin{table}
\centering
\begin{tabular}{l}
\toprule
True guide: John \\
Correct door color: blue \\
\midrule
\textit{agent goes to the wizard} \\
\textbf{Agent}: Where is the exit? \\
\textbf{Wizard}: Ask John. \\
\textit{agent goes to one guide} \\
\textbf{Agent}: Where is the exit? \\
\textbf{Jack}: Go to the red door. \\
\textit{agent goes to the other guide} \\
\textbf{Agent}: Where is the exit? \\
\textbf{John}: Go to the blue door. \\
\textit{agent goes to the blue door} \\
\textbf{Agent}: Open sesame \\
\end{tabular}
\caption{An example of a successful episode}
\label{tab:successful_episode}
\end{table}

For each episode the colors of doors and NPCs are selected randomly from a set of six and the names of the two guides are selected randomly from a set of two (Jack, John).
Furthermore, the grid width and height are randomized from the minimal size of 5 up to 8 and the NPCs and the agent are placed randomly inside (omitting locations in front of doors).


\subsubsection{Required social skills}
\label{app:talkitout_social_skills}

In this section we will discuss the TalkItOut environment in the context of social skills required of the agent.
The upcoming \benchname benchmark will contain various environments each specialized for testing different social skills i.e. some will be specialized for multimodality and others for ToM or Social games.


\textbf{Intertwined multimodality}

To solve this task the agent must use both modalities both in the action and in the observation space.
Furthermore, this multimodality is intertwined because the progression in which the modalities are used is non-trivial.
To discuss this notion further let's imagine an example of command following.
The progression of modalities here is trivial because the agent always \textit{listens} for the command first and then \textit{looks} and \textit{moves/acts} to complete the task.
Another good example is embodied question answering. Here the agent again always first \textit{listens} to the question, then \textit{looks} and \textit{moves} in the environment to finally, at the end, \textit{speak} the answer.

In our environment, however, the agent must choose which modality to use based on the current state.
And it will often be required to switch between modalities many times.
For example, to talk to an NPC the agent first \textit{looks} to find the NPC, then it \textit{moves} to the NPC, finally the agent \textit{speaks} to it and \textit{listens} to the response.
This progression is then used, if needed, for other NPCs, and finally a similar one used to go to the correct door and open it.
Furthermore, depending on the current configuration of environment, the progression can also be different.
Usually, after finding out the correct door the agent needs to \textit{look} for it and \textit{move} to it to \textit{speak} the password, but if the true guide is already next to the correct door only \textit{looking} for the door and \textit{speaking} the password is required.

\textbf{Theory of Mind}

Since the agent must be able to infer good or bad intentions of other NPCs, a basic form of ToM is needed.
Primarily, the agent needs to infer that the wizard is well-intended, wants to help, and is therefore trustworthy.
Using the inferred trust in the wizard it is possible to infer the good intentions of the true guide, and likewise the bad intentions of the false guide.

On the other hand, as the false guide chooses which false direction to give each time asked, it is also possible to infer its ill-intentions by asking him many times in the same episode and observing this inconsistency.
If an NPC gives different answers for the same question in the same episode then it is evident its intentions are bad.

\textbf{Social games}

Since social games were not the focus of this environment, and will be studied in more detail in the upcoming environments, they are present in this environment only in a simple form.
To talk with an NPC the agent needs to stand in next to it, to get an answer the agent needs to ask "where is the exit".
These simple rules (a.k.a. social conventions) are social games i.e. grammars describing the possible and impossible interactions.
It is impossible to communicate if you are far and get directions if you ask "Where is the floor".
The agent needs to be able to extract these rules and use them in relation to all the NPCs.

\subsection{Baselines details}
\label{app:cond-details}

\begin{figure}
    \centering
    \includegraphics[width=0.8\columnwidth]{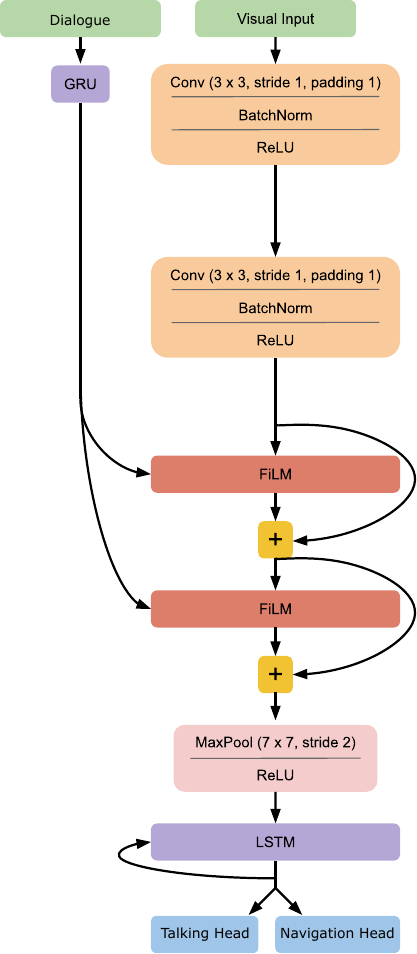}
    \caption{Our Multi-Headed BabyAI baseline DRL agent. Architecture visualization is a modified version of the one made by \citet{hui2020babyai}. We perform two modifications: 1) Instead of fixed instruction inputs our model is fed with NPC's language outputs (if the agent is near an NPC), and 2) We add a language action head, as our agent can both navigate and talk.}
    \label{fig:baby-model}
\end{figure}

\paragraph{BabyAI baseline} In this work we use a PPO-trained \cite{ppo} DRL architecture initially designed for the BabyAI benchmark \cite{chevalierboisvert2018babyai}. The policy design was improved in a follow-up paper by \citet{hui2020babyai} (more precisely, we use their \textit{original\_endpool\_res} model). See figure \ref{fig:baby-model} for a visualization of the complete architecture. First, symbolic pixel grid observations are fed into two convolutional layers \cite{lecun1989backpropagation, krizhevsky2012imagenet} (3x3 filter, stride and padding set to 1), while dialogue inputs are processed using a Gated Recurrent Unit layer \cite{gru}. The resulting image and language embeddings are combined using two FiLM attention layers \cite{film}. Max pooling is performed on the resulting combined embedding before being fed into an LSTM \cite{lstm} with a $128D$ memory vector.
The LSTM embedding is then used as input for the navigation action head, which is a two-layered fully-connected network with tanh activations and has an 8D output (i.e. 7 navigation actions and no\_op action).

In order for our agent to be able to both move and talk, we add to this architecture a talking action head, which is composed of three networks. All of them are two-layered, fully-connected networks with tanh activations, and take the LSTM's embedding as input. The first one is used as a switch: it has a one-dimensional output to choose whether the agent talks (output > 0.5) or not (output < 0.5). If the agent talks, the two other networks are used to respectively sample the template (4D output) and the word (16D output).

Note that the textual input given to the agent consists of the full dialogue history (without the "empty string" indicator) as we found it works better than giving only current utterances (see figure \ref{fig:train_no_liar}).

\begin{table}
\centering
\begin{tabular}{@{}lll@{}}
\toprule
Hyperparameter & value \\ \midrule
learning rate       & $1 e^4$    \\
GAE $\lambda$       & $0.99$    \\
clip $\epsilon$     & $1 e^5$    \\
batch size          & $1280$    \\
$\gamma$            & $0.99$   \\
recurrence          & $10$ \\
epochs              & $4$  \\
expl. bon. C        & $0.125$ \\
expl. bon. M        & $50$ \\
\bottomrule
\end{tabular}
\caption{Training hyperparametres}
\label{tab:train_hyperparmas}
\end{table}

\paragraph{Exploration bonus} 
The exploration bonus we use is inspired by recent works in intrinsically motivated exploration \cite{pathakICMl17curiosity, curiositythroughreachability, tang2017exploration}.
These intrinsic rewards estimate novelty of the currently observed state and add the novelty based bonus to the extrinsic reward. 

In this work we study a multi modal state space and we calculate the exploration bonus only on the language modality.
We count how many times was each utterance observed and compute an additional bonus based on the following equation:

\begin{equation}
r_{intr} = \frac{C}{(N(s_{lang})+1)^{M}}
\label{eq:expl_bonus}
\end{equation}

, where $M$ and $C$ are hyperparameters and $N(s_{lang})$ is the number of times the utterance $s_{lang}$ was observed during this episode.

We make our reward episodic by resetting the counts at the end of each episode.
In the current version of the environment the agent cannot hear his own utterances and the NPCs speak only when spoken to.
Therefore, this exploration bonus can be seen as analogous to social influence \cite{jacques2019-socialinfl} in the language modality, as the reward is given upon \textit{making the NPC respond}. 

Our verbal episodic intrinsic reward, which uses only the language modality, is a good example of a bias that had to be discovered for training social agents.


\section{Additional experiments}
\label{app:add-results}

In this section we will discuss some additional experiments we ran on the two environments.

\begin{figure*}
    \centering
    \begin{subfigure}[b]{.5\textwidth}
        \centering
        \includegraphics[width=\textwidth]{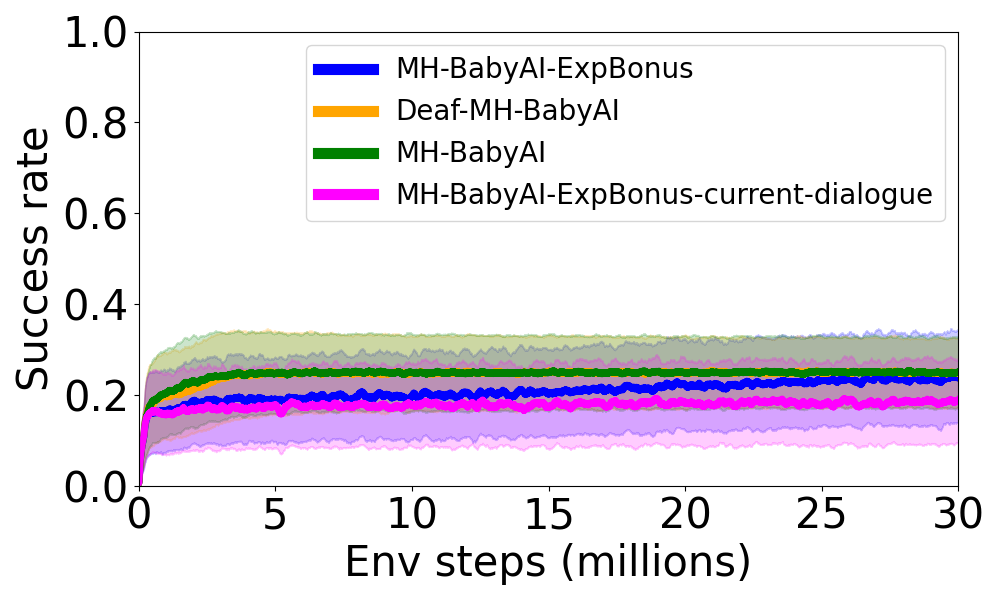}
        \caption{Full environment}
        \label{fig:train_full}
    \end{subfigure}%
    \begin{subfigure}[b]{.5\textwidth}
        \centering
        \includegraphics[width=\textwidth]{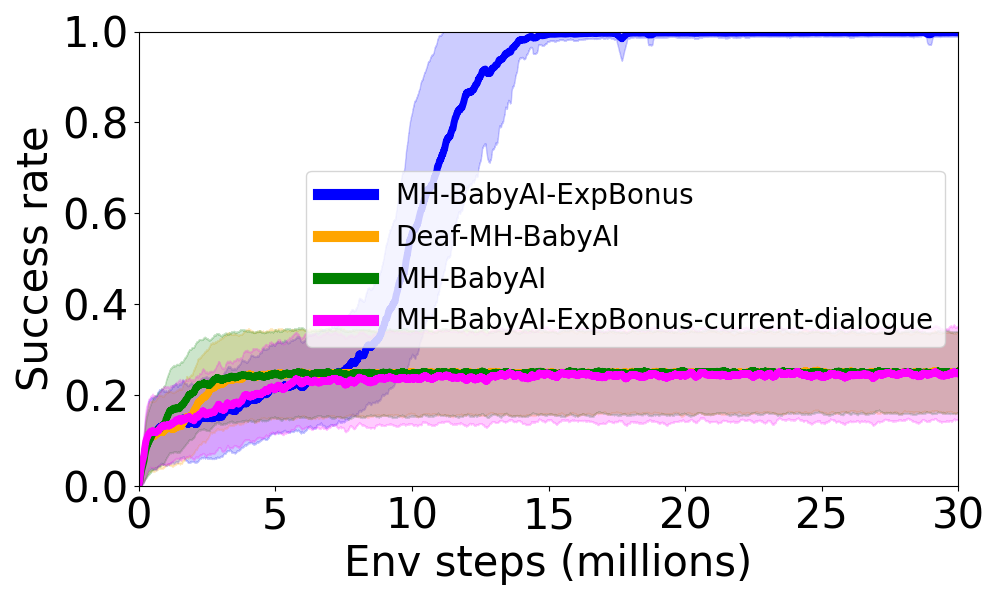}
        \caption{"NO liar NPC" environment}
        \label{fig:train_no_liar}
    \end{subfigure}%
    \caption{Training configuration experiments}
    \label{fig:train_conf}
\end{figure*}

Figure \ref{fig:train_conf} shows the success rates for the configurations discussed in the main text and displayed in table \ref{tab:results}.
One additional configuration shown in this figure is the one denoted "MH-BabyAI-ExpBonus-current-dialogue".
In this configuration instead of giving the agent the full dialogue history only the dialogue observed in the current timestep or, if no dialogue is observed, an "empty" indicator string ("NA") is given. It is clearly visible that this configuration is inferior to the one providing the agent with the full dialogue history.

\begin{figure*}
    \centering
    \begin{subfigure}[b]{.5\textwidth}
        \centering
        \includegraphics[width=\textwidth]{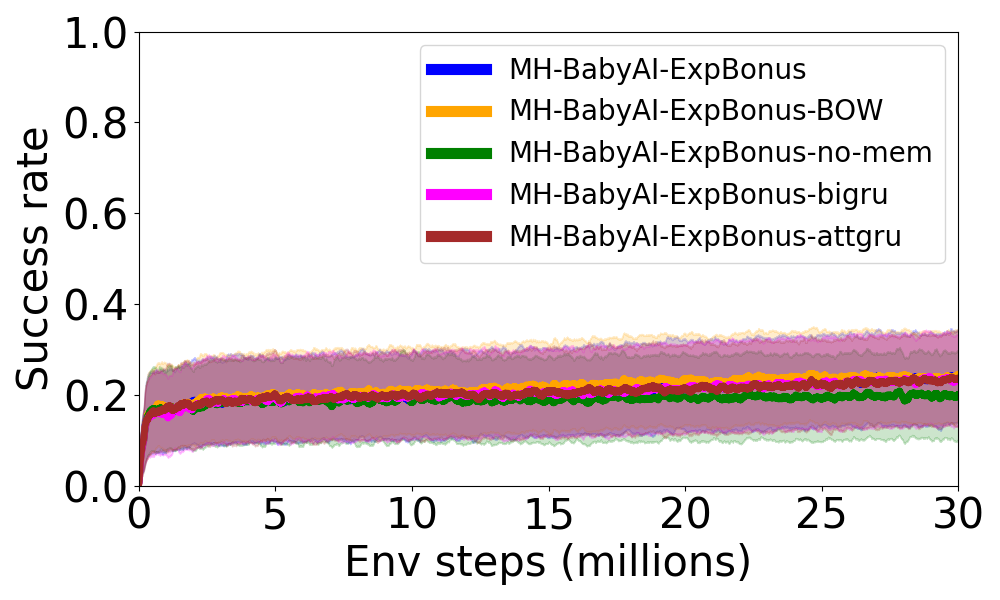}
        \caption{Full environment}
        \label{fig:arch_full}
    \end{subfigure}%
    \begin{subfigure}[b]{.5\textwidth}
        \includegraphics[width=\textwidth]{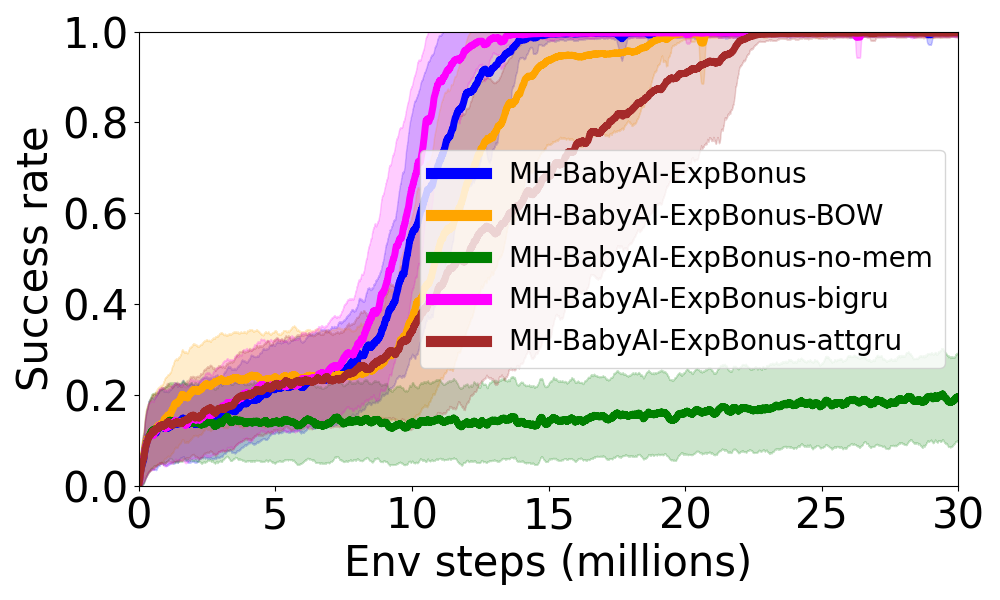}
        \caption{"NO liar NPC" environment}
        \label{fig:arch_no_liar}
    \end{subfigure}%
    \caption{Architectural experiments}
    \label{fig:arch_conf}
\end{figure*}

Furthermore, we ran some experiments varying the architecture of the network. These experiments are visible in figure \ref{fig:arch_conf}.
In this figure the "no-mem" refers to the network lacking the final LSTM layer. However, this network still has some form of memory as the full dialogue history is fed into the GRU unit.
We can see that, without the LSTM, the agent is not able to solve the ablation environment.
We likewise ran experiments where we replaced the GRU unit with a bidirectional-GRU unit ("bigru"), and where we used attention on top of that GRU unit ("attgru")\footnote{The attention vector was computed using a linear layer on the LSTM's hidden state from the previous step.}. 
We also experimented with a different approach of representing the vision modality: a BOW based embedding as used in \citep{hui2020babyai}.
In this approach each grid is represented as a BOW and this representaiton used to retrieve the embedding from a trainable lookup table.
We can see that these architectures, like the basic one with the GRU, are able to solve the environment without the liar NPC, but not the full environment.


\end{document}